\pdfoutput=1

\documentclass[11pt]{article}

\usepackage[preprint]{acl}

\usepackage{times}
\usepackage{latexsym}

\usepackage[T1]{fontenc}

\usepackage[utf8]{inputenc}

\usepackage{microtype}

\usepackage{inconsolata}

\usepackage{graphicx}
\usepackage{amsmath}
\usepackage{amssymb}
\usepackage{bbm}
\usepackage{xcolor}
\usepackage{color}
\usepackage{float}
\usepackage{booktabs}
\usepackage{multirow}
\usepackage{makecell}

\usepackage{colortbl}
\usepackage{pifont}
\usepackage{graphicx}
\usepackage{subcaption}
\usepackage{algorithm}
\usepackage{algorithmicx}
\usepackage{algpseudocode}
\usepackage{fdsymbol}
\usepackage{pdfpages}

\usepackage{listings}
\usepackage{xcolor}

\usepackage{tcolorbox}
\tcbuselibrary{listings}
\usepackage{cuted} 

\usepackage{CJKutf8}

\definecolor{orange}{HTML}{FF9933}
\definecolor{light-blue}{HTML}{3399FF}
\definecolor{dark-blue}{HTML}{3333FF}
\title{From Synthetic to Native: Benchmarking Multilingual Intent Classification in Logistics Customer Service}

\author{Haoyu He$^{1,2}$ \thanks{This work was mainly done while the author was an intern at J\& T Express.}  Jinyu Zhuang$^{1}$  Haoran Chu$^{1,2}$\footnotemark[1] \\Shuhang Yu$^{1}$  J\& T AI Group$^{1}$ \\ Hao Wang$^{2}$ Kunpeng Han$^{1}$ \thanks{~Corresponding author}\\
  J\&T Express$^1$\\
  School of Information Science and Technology, ShanghaiTech University$^2$ \\
    {\tt \{robert.zhuang,wayne.yu,lion.han\}@jtexpress.com},\\ 
    \{hehy12024,chuhr2023,wanghao1\}@shanghaitech.edu.cn
 }

\begin{document}
\maketitle
\begin{abstract}
Multilingual intent classification is central to customer-service systems on global logistics platforms, where models must process noisy user queries across languages and hierarchical label spaces. Yet most existing multilingual benchmarks rely on machine-translated text, which is typically cleaner and more standardized than native customer requests and can therefore overestimate real-world robustness. We present a public benchmark for hierarchical multilingual intent classification constructed from real logistics customer-service logs. The dataset contains approximately 30K de-identified, stand-alone user queries curated from 600K historical records through filtering, LLM-assisted quality control, and human verification, and is organized into a two-level taxonomy with 13 parent and 17 leaf intents. English, Spanish, and Arabic are included as seen languages, while Indonesian, Chinese, and additional test-only languages support zero-shot evaluation. To directly measure the gap between synthetic and real evaluation, we provide paired native and machine-translated test sets and benchmark multilingual encoders, embedding models, and small language models under flat and hierarchical protocols. Results show that translated test sets substantially overestimate performance on noisy native queries, especially for long-tail intents and cross-lingual transfer, underscoring the need for more realistic multilingual intent benchmarks.\footnote{Current resources are available at \url{https://anonymous.4open.science/r/MICCS}, and will be officially released on Hugging Face following the OpenReview process.}
\end{abstract}

\section{Introduction}

Intent classification is a core problem in natural language understanding and underpins many interactive systems, including customer-service automation, virtual assistants, and tool-using language models \citep{larson-etal-2019-evaluation,casanueva-etal-2020-efficient,schick2023toolformer}. In multilingual customer-service settings, the task is particularly challenging: semantically equivalent intents may be expressed quite differently across languages and dialects, while real user queries are often short, noisy, and domain-specific. On global logistics platforms, systems must route requests about tracking, delivery issues, address changes, refunds, and claims, often under hierarchical label taxonomies \citep{schuster-etal-2019-cross-lingual,gung-etal-2023-natcs,plaud-etal-2024-revisiting}.

Although existing multilingual benchmarks have enabled substantial progress, they only partly reflect these deployment conditions. Widely used resources such as MTOP and MASSIVE provide broad language coverage, but much of their multilingual data comes from machine translation or controlled data collection rather than naturally occurring customer requests \citep{li-etal-2021-mtop,fitzgerald-etal-2023-massive}. Other customer-service datasets are more realistic in domain, but often focus on multi-turn dialogue, sentiment, search, or matching tasks rather than stand-alone intent-routing queries with hierarchical labels \citep{chen-etal-2021-action,gung-etal-2023-natcs,wang-etal-2025-ecom-bench}. As a result, current benchmarks may underrepresent the lexical variability, shorthand, code-mixing, and long-tail label distributions that characterize production traffic \citep{artetxe-etal-2023-revisiting,dutta-chowdhury-etal-2022-towards}.

This mismatch also affects evaluation. Machine-translated test sets are typically cleaner and more standardized than native user text, and may therefore yield overly optimistic estimates of multilingual robustness \citep{artetxe-etal-2023-revisiting,dutta-chowdhury-etal-2022-towards}. Measuring this synthetic-to-native gap requires matched evaluation settings in which the underlying intent is fixed while the surface form differs between native and translated queries.

To address this gap, we introduce a public benchmark for hierarchical multilingual intent classification built from real logistics customer-service logs. The dataset contains approximately 30K de-identified labeled queries curated from roughly 600K historical records through rule-based filtering, LLM-assisted quality control, and human verification. It is organized into a two-level taxonomy with 13 parent intents and 17 leaf intents. English, Spanish, and Arabic are included as seen languages, while Indonesian, Chinese, and additional test-only languages support zero-shot evaluation. Because the benchmark is derived from native, stand-alone user queries used for routing, it preserves the noise and distributional properties of real customer-service traffic.

To explicitly study the gap between synthetic and real multilingual evaluation, we additionally provide paired native and machine-translated test sets. We benchmark multilingual encoders, embedding models, and small language models under both flat and hierarchical protocols. Across model families, translated evaluation consistently yields higher performance than native evaluation, with especially large gaps for long-tail intents and cross-lingual transfer. These results suggest that translated benchmarks can substantially overstate robustness on real multilingual user traffic.

In summary, our contributions are as follows:
\begin{itemize}
    \item We release a public real-world benchmark for hierarchical multilingual intent classification in logistics customer service, together with a practical curation pipeline that combines de-identification, filtering, LLM-assisted quality control, and human verification.
    \item We provide paired native and machine-translated test sets for direct, controlled measurement of the synthetic-to-native evaluation gap.
    \item We benchmark multilingual encoders, embedding models, and small language models under flat and hierarchical evaluation settings, with a focus on long-tail prediction, noisy native text, and cross-lingual generalization.
    \item We show empirically that machine-translated evaluation can systematically overestimate model performance relative to native-query evaluation.
\end{itemize}

\section{Dataset Construction}
\label{sec:construction}

We construct a multilingual intent benchmark from historical customer-service logs. The released data consist of de-identified user queries paired with a practical two-level intent taxonomy.

\begin{figure*}[t]
    \centering
    \fbox{\parbox{1\linewidth}{
    \centering
    \includegraphics[width=\textwidth]{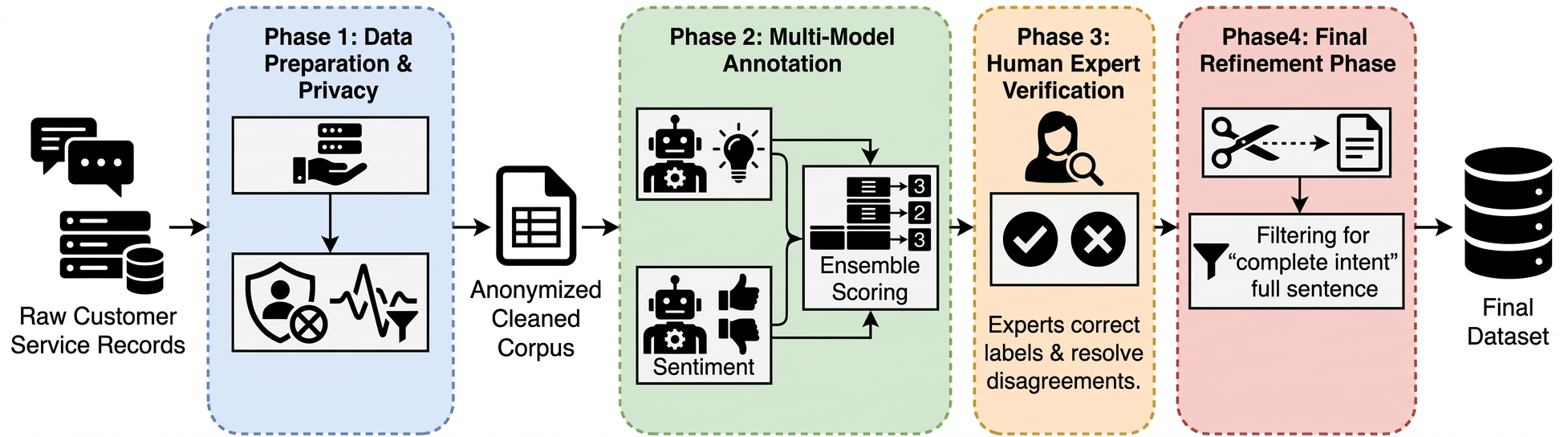}
    }}
    \caption{Construction pipeline for the benchmark.}
    \label{fig:pipeline}
\end{figure*}

\subsection{Construction Pipeline}

\paragraph{Source data.}
The raw pool comprises historical customer-service interactions. From this pool, we extract user utterances that express identifiable service intents. The released benchmark contains approximately 30K labeled queries. English, Spanish, and Arabic are used as seen languages, while Indonesian, Chinese, and additional held-out languages are reserved for zero-shot transfer evaluation.

\paragraph{De-identification and cleaning.}
Before annotation and release, we remove personally identifiable information, such as tracking numbers and detailed addresses, using rule-based filters followed by manual checks. We then exclude bot-generated templates, exact and near duplicates, malformed queries, and utterances without clear intent. Deduplication combines string matching with embedding-based semantic similarity. This process reduces annotation noise while preserving naturally occurring linguistic variation.

\paragraph{Taxonomy construction.}
We group the cleaned queries by operational function and user goal, and iteratively refine the label set to balance coverage, label separability, and per-class support. The final taxonomy is organized as a two-level hierarchy with 13 parent intents and 17 leaf intents. Details of the clustering configuration are provided in Appendix~\ref{sec:clustering-analysis} and Figure~\ref{fig:kmeans}.

\paragraph{Annotation and quality control.}
We use a semi-automatic annotation pipeline with human verification. Candidate samples are pre-screened with LLM-assisted filtering and consistency checks, and are then reviewed by two annotators for each language. A 5\% subset is double-annotated for test-set construction; annotator agreement on this subset is 95\%. Remaining disagreements are adjudicated by a third expert, and samples without consensus are removed.

\section{Dataset Characteristics}
\label{sec:characteristics}

\subsection{Overview}

Appendix Table~\ref{tab:dataset-overview} summarizes the corpus statistics. The release contains approximately 30K queries, split into train/dev/test at a ratio of 20:1:1. A representative de-identified example is shown below:

\begin{tcblisting}{
    colback=gray!10, 
    colframe=gray!50, 
    arc=4pt, 
    listing only, 
    listing options={
        basicstyle=\ttfamily\small, 
        breaklines=true, 
        columns=fullflexible
    }
}
text: Where is my parcel? It has not moved for three days. 
class: Shipment Tracking / Delay
label: i1
\end{tcblisting}

To improve label generation stability in generative classification, we prepend a fixed prefix (e.g., ``i1'') to all label IDs \citep{xiao-etal-2023-streaming-llm}.

\subsection{Native vs.\ Synthetic Evaluation}

A key feature of the benchmark is the distinction between native and synthetic multilingual evaluation. Native evaluation uses a 674-example human-verified subset of held-out multilingual traffic, with cross-language semantic duplicates removed. 

For synthetic evaluation, we construct paired machine-translated test sets from a reliability-filtered source subset and verify the translations manually. These translated sets preserve intent semantics while removing much of the noise present in native traffic. We therefore treat them as a controlled synthetic baseline rather than a direct proxy for deployment conditions. Comparing performance across the two settings allows us to measure the gap between translated and native evaluation.

\subsection{Hierarchical Labels, Long-Tail Sparsity, and Noise}

The benchmark supports both a 13-way parent classification task and a 17-way leaf classification task (Appendix Table~\ref{tab:taxonomy}). The leaf setting is more difficult because it requires finer operational distinctions and exhibits greater class imbalance.

\begin{figure}[t]
    \centering
    \includegraphics[width=0.4\textwidth]{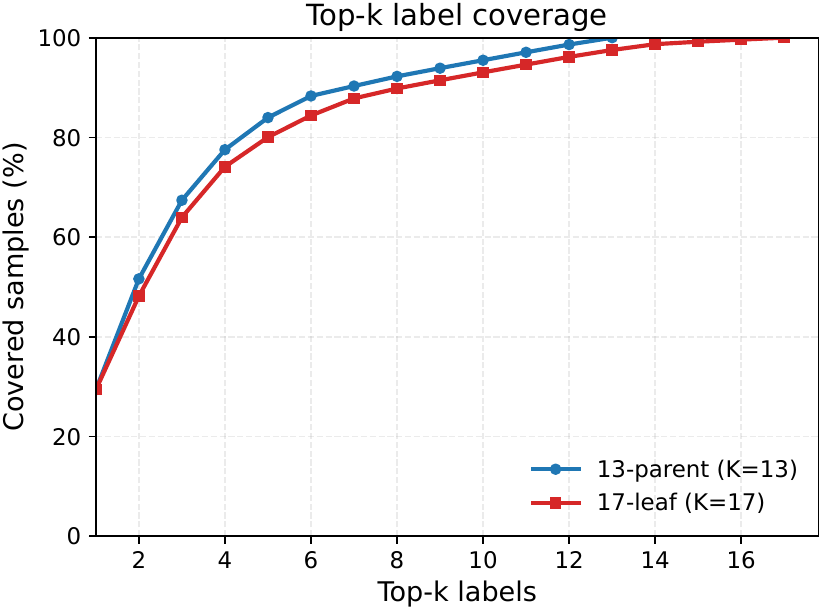}
    \caption{Top-$k$ label coverage for the parent and leaf tasks. The 17-way leaf setting is substantially more long-tailed than the 13-way parent setting.}
    \label{fig:longtail}
\end{figure}

As shown in Figure~\ref{fig:longtail}, the 17-way leaf task is substantially more long-tailed than the parent task, consistent with production traffic in which a small number of intents dominate and many low-frequency cases remain sparse. The native queries also retain naturally occurring noise, including shorthand, misspellings, colloquial phrasing, and cross-lingual variation, which are often absent from translated benchmarks.

\section{Experiments}
\subsection{Benchmark Tasks and Evaluation Protocol}
\label{sec:tasks}

We evaluate multilingual intent understanding under three task settings designed to reflect common customer-service routing needs. The first is a 13-way parent-intent classification task, which captures coarse-grained operational routing such as distinguishing tracking-related requests from fee, claim, or invoice issues. The second is a 17-way flat leaf-intent classification task, where models directly predict one of the 17 leaf intents without using the label hierarchy at inference time. This is our main fine-grained benchmark setting, as it requires models to disambiguate operationally similar intents with overlapping surface forms. The third is a 17-way hierarchical classification task, in which models predict leaf intents under the released parent--child taxonomy. This hierarchy is deployment-oriented rather than ontology-deep: in production, the system operates over the 17-way operational label space, after which business routing rules merge outputs into parent categories. As a result, some parent nodes currently map to a single leaf, while others group multiple leaves. We use this setting primarily to test whether hierarchy-aware decoding improves accuracy under practical routing constraints and to provide a controlled inference interface for generative models.

Unless otherwise stated, all trainable models are trained on the released training split and selected on the development split using only the seen benchmark languages, namely English, Spanish, and Arabic. Closed-source LLMs, which cannot be fine-tuned in the same way, are evaluated in a few-shot prompting setting. All results are reported under the benchmark's natural traffic distribution, without class rebalancing or synthetic augmentation.

For unseen test-only languages such as Indonesian and Chinese, we evaluate zero-shot transfer without target-language supervision, translated training data, or target-language prompt tuning. To study the effect of translated versus native evaluation within the same target language, we compare model performance under two conditions: (i) machine-translated test queries, translated using TranslateGemma-27B~\citep{finkelstein2026translategemma}, and (ii) native user queries. We further construct a business-rewritten \emph{native-style} variant, in which domain operators rewrite translated outputs into naturally used operational phrasing while preserving the original intent labels. This setup isolates distributional differences between translated and native evaluation while holding the target language fixed.

Our primary metric is \textbf{Accuracy/Micro-F1}. This choice is intentional: the benchmark preserves the naturally imbalanced traffic distribution of real customer-service systems, and overall routing accuracy is the most deployment-relevant objective in this setting. We report Macro-F1 in Appendix~\ref{sec:macro} for long-tail and class-balance analysis. All hyperparameters, prompts, and checkpoints are selected on the development split only. Test sets, including translated and unseen-language test sets, are used strictly for final evaluation.

\subsection{Baselines}
\label{sec:baselines}

We compare four baseline families chosen to reflect common production trade-offs among latency, cost, and multilingual generalization. To stay close to realistic serving constraints, we focus primarily on models below 1B parameters, while also including a small set of API-based closed-source LLMs for reference.

For multilingual encoder baselines, we evaluate \textbf{mBERT}~\citep{devlin-etal-2019-bert}, \textbf{XLM-R}~\citep{conneau-etal-2020-unsupervised}, and an internal optimized variant, \textbf{\textsc{mmBERT}}~\citep{marone-etal-2025-mmbert}. These models use a fixed output label space and remain strong low-latency baselines for intent routing.

For embedding-based classification, we evaluate \textbf{EmbeddingGemma}~\citep{lee2025embeddinggemma} paired with a lightweight classifier, as well as a \textbf{Gemma 3 270M (embedding+classifier)} variant~\citep{gemma2024}. This setup is particularly relevant for organizations that already maintain shared multilingual embedding infrastructure.

We further evaluate instruction-tuned small language models, including \textbf{Gemma 3 270M}, \textbf{Gemma 3 1B}~\citep{gemma2024}, and \textbf{Qwen 3 0.6B}~\citep{qwen3-technical-report}. These models are attractive in practice because they support zero-shot transfer and lightweight taxonomy updates without requiring retraining of a large encoder.

Finally, we compare against three prompted closed-source LLMs accessed via API: \textbf{Seed-1.8}~\citep{seed-1-8-modelark}, \textbf{DeepSeek V3.2}~\citep{deepseek-v3-technical-report}, and \textbf{Qwen-Max-Preview}~\citep{qwen-api-model-studio}.

All baselines use the same label inventory, train/dev/test splits, and evaluation scripts.
The training settings can be found in Appendx~\ref{app:tp}.
\section{Results}
\label{sec:results}

\subsection{Seen-Language Supervised Performance}

Table~\ref{tab:seen-results} reports supervised results on the seen benchmark languages (English, Spanish, and Arabic); following Section~\ref{sec:tasks}, we report Accuracy/Micro-F1 as the primary metric.

Across model families, the 13-way parent task is consistently easier than the 17-way leaf tasks, reflecting the added difficulty of separating operationally adjacent leaf intents under the benchmark's natural long-tail distribution. Additional Macro-F1 results are provided in Appendix~\ref{sec:macro}.

\begin{table}[t]
\centering
\caption{Seen-language supervised benchmark results on English, Spanish, and Arabic (Accuracy/Micro-F1, \%).}
\label{tab:seen-results}
\resizebox{\linewidth}{!}{%
\begin{tabular}{lccc}
\toprule
\textbf{Model} & \textbf{13-way} & \textbf{17-flat} & \textbf{17-hier} \\
\midrule
\multicolumn{4}{l}{\textit{Multilingual encoders}} \\
mBERT & 92.62 & 91.38 & N/A \\
XLM-R & 94.00 & 91.85 & N/A \\
\textsc{mmBERT} & 94.69 & 93.15 & N/A \\
\midrule
\multicolumn{4}{l}{\textit{Embedding-based classifiers}} \\
EmbeddingGemma  & 94.85 & 93.38  & N/A \\
Gemma 3 270M (embedding+classifier) & 93.62 & 92.15 & N/A \\
\midrule
\multicolumn{4}{l}{\textit{Instruction-tuned SLMs}} \\
Gemma 3 270M & 94.08 & 93.08 & 93.23 \\
Qwen 3 0.6B & 94.23 & 93.46 & 93.77 \\
Gemma 3 1B & 95.00 & 93.92 & 94.08  \\
\midrule
\multicolumn{4}{l}{\textit{closed-source LLMs}} \\
Seed-1.8  & 91.69 & 92.23  &  91.23  \\
DeepSeek V3.2 & 93.92 & 90.92 & 89.85 \\
Qwen-Max-Preview & 94.77 & 91.15 &  92.92  \\
\bottomrule
\end{tabular}%
}
\end{table}

Detailed Macro-F1 results are provided in Table~\ref{tab:appendix-seen-f1}.

\subsection{Real-World Robustness: Synthetic-to-Native Gap and Zero-Shot Transfer}

A central goal of the benchmark is to measure performance under realistic multilingual deployment conditions. Beyond supervised evaluation on seen languages, we therefore examine two additional aspects of robustness: the gap between translated and native user queries, and zero-shot transfer to unseen languages. Table~\ref{tab:synthetic-native-results} summarizes both analyses.

\paragraph{Synthetic-to-Native Robustness Gap.}
Real customer utterances often contain shorthand, misspellings, colloquialisms, and other naturally occurring noise that translation pipelines tend to smooth away. To isolate this effect, we compare model performance on translated test sets and native user queries within the \emph{same target language}. Because the target language is held fixed, the resulting gap reflects distributional differences rather than language identity. In our experiments, translated performance is higher in most settings. This trend likely reflects both translation-induced normalization and the construction of the synthetic set, which pre-filters source items to those for which GPT and Gemini are both correct and mutually consistent before translation.

\begin{CJK*}{UTF8}{gbsn}
For concreteness, we show one de-identified example from the dataset illustrating this smoothing effect:

\begin{tcolorbox}[
    colback=gray!10,
    colframe=gray!50,
    arc=4pt
]
\ttfamily\small
original: Do you have an API for automating shipments?\\
translated: 你们是否提供 API，以便实现邮件发送的自动化？\\
native: 你们有能自动发快递的接口吗？
\end{tcolorbox}
\end{CJK*}

\paragraph{Zero-Shot Transfer to Unseen Languages.}
We additionally report zero-shot transfer results for 17-way flat classification on unseen test-only languages, including Indonesian and Chinese. In this setting, models are trained only on English, Spanish, and Arabic, and are evaluated directly on unseen-language test sets without target-language supervision. This setup reflects a realistic deployment scenario in which systems must support new markets before localized labeled data becomes available, and it helps identify which model families transfer most effectively.

\begin{table*}[t]
\centering
\caption{Synthetic-to-native robustness gap under same-language evaluation (Accuracy/Micro-F1, \%). Models are evaluated on translated (Trans.) versus native user queries across varying task granularities. \textbf{Bold} values indicate the best performance within each architectural category (Masked LMs vs. Causal LLMs). \underline{Underlined} values with $\uparrow$ explicitly highlight instances where the lightweight \textbf{Gemma 3 270M} outperforms the similarly-sized, classification-specialized \textbf{mmBERT}. Notably, Causal LLMs support the 17-way hierarchical (Hier) formulation natively, whereas implementing hierarchy-aware constrained decoding for standard BERT-style classifiers is typically more involved than flat classification.}
\label{tab:synthetic-native-results}
\begin{tabular}{ll cccccc}
\toprule
\multirow{2}{*}{\textbf{Model}} & \multirow{2}{*}{\textbf{Lang.}} & \multicolumn{2}{c}{\textbf{13-way Parent}} & \multicolumn{2}{c}{\textbf{17-way Flat}} & \multicolumn{2}{c}{\textbf{17-way Hier}} \\
\cmidrule(lr){3-4} \cmidrule(lr){5-6} \cmidrule(lr){7-8}
& & \textbf{Trans.} & \textbf{Native} & \textbf{Trans.} & \textbf{Native} & \textbf{Trans.} & \textbf{Native} \\
\midrule
\multirow{3}{*}{mBERT} 
& ZH & 64.84 & 68.10 & 68.40 & 65.13 & N/A & N/A \\
& ID & 54.60 & 53.41 & 52.82 & 51.63 & N/A & N/A \\
& TH & 19.29 & 17.80 & 20.03 & 18.84 & N/A & N/A \\
\addlinespace 
\multirow{3}{*}{\textsc{XLM-R}} 
& ZH & 90.36 & 87.54 & 88.43 & 85.61 & N/A & N/A \\
& ID & 92.58 & 90.36 & 90.80 & 88.72 & N/A & N/A \\
& TH & \textbf{87.83} & \textbf{89.32} & 82.94 & 82.64 & N/A & N/A \\
\addlinespace
\multirow{3}{*}{\textsc{mmBERT}} 
& ZH & \textbf{90.95} & \textbf{89.32} & \textbf{89.91} & \textbf{87.54} & N/A & N/A \\
& ID & \textbf{93.60} & \textbf{91.99} & \textbf{91.69} & \textbf{89.91} & N/A & N/A \\
& TH & 87.54 & 85.31 & \textbf{84.27} & \textbf{83.83} & N/A & N/A \\
\midrule\midrule 
\multirow{3}{*}{\textbf{Gemma 3 270M}} 
& ZH & \underline{91.69}$^\uparrow$ & \underline{90.06}$^\uparrow$ & \underline{90.95}$^\uparrow$ & \underline{89.02}$^\uparrow$ & 88.58 & 89.91 \\
& ID  & 90.06 & 88.43 & 88.58 & 86.94 & 89.32 & 88.58 \\
& TH & 86.65 & 83.38 & \underline{85.61}$^\uparrow$ & \underline{84.87}$^\uparrow$ & 85.76 & 83.09 \\
\addlinespace
\multirow{3}{*}{Qwen 3 0.6B} 
& ZH & 93.02 & 92.14 & 92.58 & 91.69 & 92.14 & 91.99 \\
& ID & 88.72 & 87.53 & 86.65 & 85.01 & 86.35 & 85.31 \\
& TH & 85.31 & 84.87 & 85.61 & 84.72 & 85.31 & 84.57 \\
\addlinespace
\multirow{3}{*}{Gemma 3 1B} 
& ZH & \textbf{95.25} & \textbf{93.47} & \textbf{94.81} & \textbf{93.32} & \textbf{94.96} & \textbf{92.73} \\
& ID & \textbf{94.51} & \textbf{93.32} & \textbf{91.39} & \textbf{89.91} & \textbf{93.62} & \textbf{92.28} \\
& TH & \textbf{89.61} & \textbf{87.69} & \textbf{87.98} & \textbf{86.35} & \textbf{89.61} & \textbf{87.98} \\
\bottomrule
\end{tabular}%
\end{table*}

\paragraph{Architectural Shifts and Scaling Laws.} 
Traditionally, Masked Language Models (MLMs) such as XLM-R and mmBERT have been strong baselines for NLU classification tasks. Our results suggest a meaningful shift: despite being a generative Causal LM, the lightweight Gemma 3 270M outperforms the classification-specialized mmBERT (307M parameters) in multiple settings, including all Chinese evaluation conditions and the Thai 17-way flat task in our benchmark. This indicates that modern Small Language Models (SLMs) can be highly competitive with optimized BERT-style classifiers for intent classification. The results also align with expected scaling behavior: Gemma 3 1B consistently performs best across languages, data conditions, and task formulations in our experiments, outperforming both Gemma 3 270M and Qwen 3 0.6B.

\paragraph{Robustness to Granularity and Production Deployment.} 
A critical operational requirement in real-world systems is the ability to handle fine-grained, complex user intents. As shown in Table~\ref{tab:synthetic-native-results}, when transitioning from 13-way parent categories to 17-way flat classification, BERT-style models generally show larger performance drops than Gemma models in our benchmark. In addition, causal LLMs can naturally express 17-way hierarchical predictions, while BERT-style classifiers usually require additional design (e.g., constrained decoding or multi-stage heads) to model hierarchy consistently. Considering the observed native-query robustness, performance under finer-grained taxonomies, and cross-lingual transfer results, we have progressively shifted our online system from legacy BERT-based pipelines toward Gemma-based models.


\section{Conclusion}

We have presented a public real-world benchmark for hierarchical multilingual intent classification constructed from logistics customer-service logs. The benchmark contains approximately 30K de-identified labeled queries organized into a two-level taxonomy with 13 parent and 17 leaf intents, covering three seen languages (English, Spanish, Arabic) and multiple unseen test-only languages (Indonesian, Chinese, Thai) for zero-shot evaluation. A key design principle is the inclusion of paired native and machine-translated test sets, which enables direct measurement of the synthetic-to-native evaluation gap.

Our experiments across multilingual encoders, embedding-based classifiers, and instruction-tuned small language models show that machine-translated evaluation often overestimates robustness relative to native user-query evaluation; for the detailed breakdown, see the synthetic-to-native results table (Table~\ref{tab:synthetic-native-results}). We also find that fine-tuned small models are generally more reliable than prompted API LLMs for fine-grained intent routing, while hierarchical constraints provide modest gains in some settings.

We hope this benchmark serves as a practical resource for developing and evaluating multilingual intent systems under realistic conditions—including long-tail label distributions, noisy native text, and genuine cross-lingual transfer—and encourages the community to complement translated evaluations with native-language benchmarks when assessing real-world multilingual robustness.

\section*{Limitations}
Our benchmark provides a strong starting point for realistic multilingual intent evaluation, while also highlighting several promising directions for future work. Although the current release covers six languages, extending it to more low-resource languages and regions would further improve global representativeness; similarly, incorporating data from multiple service providers could strengthen cross-company and cross-domain generalization. The present taxonomy (13 parent and 17 leaf intents) is intentionally designed for practical routing, and future iterations can further enrich it with deeper hierarchies and multi-label relations to support more complex reasoning. For synthetic evaluation, we use one translation system and pre-filtered source examples to keep comparisons controlled; an interesting next step is to include multiple translation engines and broader sampling strategies to better disentangle translation and filtering effects. Our current evaluation emphasizes single-utterance intent classification with accuracy as the primary metric, and can be naturally complemented by future studies on multi-turn dialogue, slot filling, end-to-end task success, richer error analysis, and statistical significance testing. Finally, because the benchmark reflects customer-service traffic from a specific period, continual updates will be valuable for studying temporal and distributional drift in evolving production settings.

\section*{Acknowledgement}

We thank the annotation team and domain experts at J\&T Express for their contributions to data curation, taxonomy design, and quality verification. We also thank the anonymous reviewers for their constructive feedback. This work was supported in part by the J\&T AI Group and ShanghaiTech University.


\bibliography{custom}

\appendix
\section{dataset supplement}

This section complements the dataset-construction and task-definition details in Sec.~\ref{sec:construction} and Sec.~\ref{sec:tasks}. We provide consolidated release metadata, split settings, and hierarchy mapping so that the evaluation setup in the main paper can be reproduced without ambiguity.

\begin{table*}[t]
\centering
\caption{Benchmark overview. Precise per-language counts can be inserted once the final release manifest is fixed.}
\label{tab:dataset-overview}
\begin{tabular}{lp{0.62\linewidth}}
\toprule
\textbf{Field} & \textbf{Value} \\
\midrule
Raw source records & $\sim$600K historical human customer-service records \\
Released real labeled queries & $\sim$30K \\
Primary split ratio & train/dev/test = 20:1:1 \\
Seen benchmark languages & English, Spanish, Arabic \\
Unseen test-only languages & Indonesian, Chinese, and additional held-out languages \\
Native multilingual evaluation set & 1,300 examples, with ES:EN:AR = 2:1:1 \\
Synthetic evaluation & Machine-translated multilingual test sets \\
Translation source language & Spanish \\
Translation system & TranslateGemma-27B \\
Public release & De-identified utterances, labels, language tags, taxonomy, splits, translated test sets, evaluation scripts \\
\bottomrule
\end{tabular}
\end{table*}

Table~\ref{tab:dataset-overview} summarizes the same core setup used in the main experiments, including language coverage, native-versus-synthetic evaluation design, and released artifacts.

\begin{table*}[t]
\centering
\caption{Classification label names and leaf-to-parent mapping for the released hierarchy.}
\label{tab:taxonomy}
\small
\begin{tabular}{clcl}
\toprule
\textbf{Leaf ID} & \textbf{Leaf intent} & \textbf{Parent ID} & \textbf{Parent intent} \\
\midrule
0  & Human Agent                                   & 0  & Human Agent \\
1  & Shipment Tracking / Delay                     & 1  & Shipment Tracking / Delay \\
2  & Delivery Failure / Status Dispute             & 2  & Delivery Failure / Status Dispute \\
3  & Change Delivery Details / Method / Time       & 3  & Delivery Change / Intercept / Return \\
4  & Issue Report (No Compensation Request)        & 4  & Damage / Missing / Confirmed Loss \\
5  & Service Scope / Transit Time / Working Hours  & 5  & Service Scope / Transit / Branch \\
6  & Service Complaint                             & 6  & Service Complaint \\
7  & Fee / Price Inquiry (No Dispute)              & 7  & Fees / Price / Dispute \\
8  & Pickup Booking                                & 8  & Pickup Booking \\
9  & Shipment Details                              & 9  & Shipment Details \\
10 & Packing / Restrictions / Insurance            & 10 & Packing / Restrictions / Insurance \\
11 & Invoice Request                               & 11 & Invoice Request \\
12 & Business Partnership                          & 12 & Business Partnership \\
13 & Intercept / Return Shipment                   & 3  & Delivery Change / Intercept / Return \\
14 & Compensation / Claim / Claim Progress         & 4  & Damage / Missing / Confirmed Loss \\
15 & Branch / Service Point Existence or Location  & 5  & Service Scope / Transit / Branch \\
16 & Fee Dispute / Incorrect Charge                & 7  & Fees / Price / Dispute \\
\bottomrule
\end{tabular}
\end{table*}

Table~\ref{tab:taxonomy} provides the complete leaf-to-parent mapping used by the 13-way parent task and the 17-way flat/hierarchical tasks discussed in Sec.~\ref{sec:results}.

\section{Clustering Analysis}
\label{sec:clustering-analysis}

This section provides the full clustering-selection evidence corresponding to the brief summary in the main text, and explains why \(K=13\) is adopted for downstream taxonomy construction.

\begin{figure*}[t]
    \centering
     \begin{minipage}{0.49\textwidth}
        \centering
        \includegraphics[width=\linewidth]{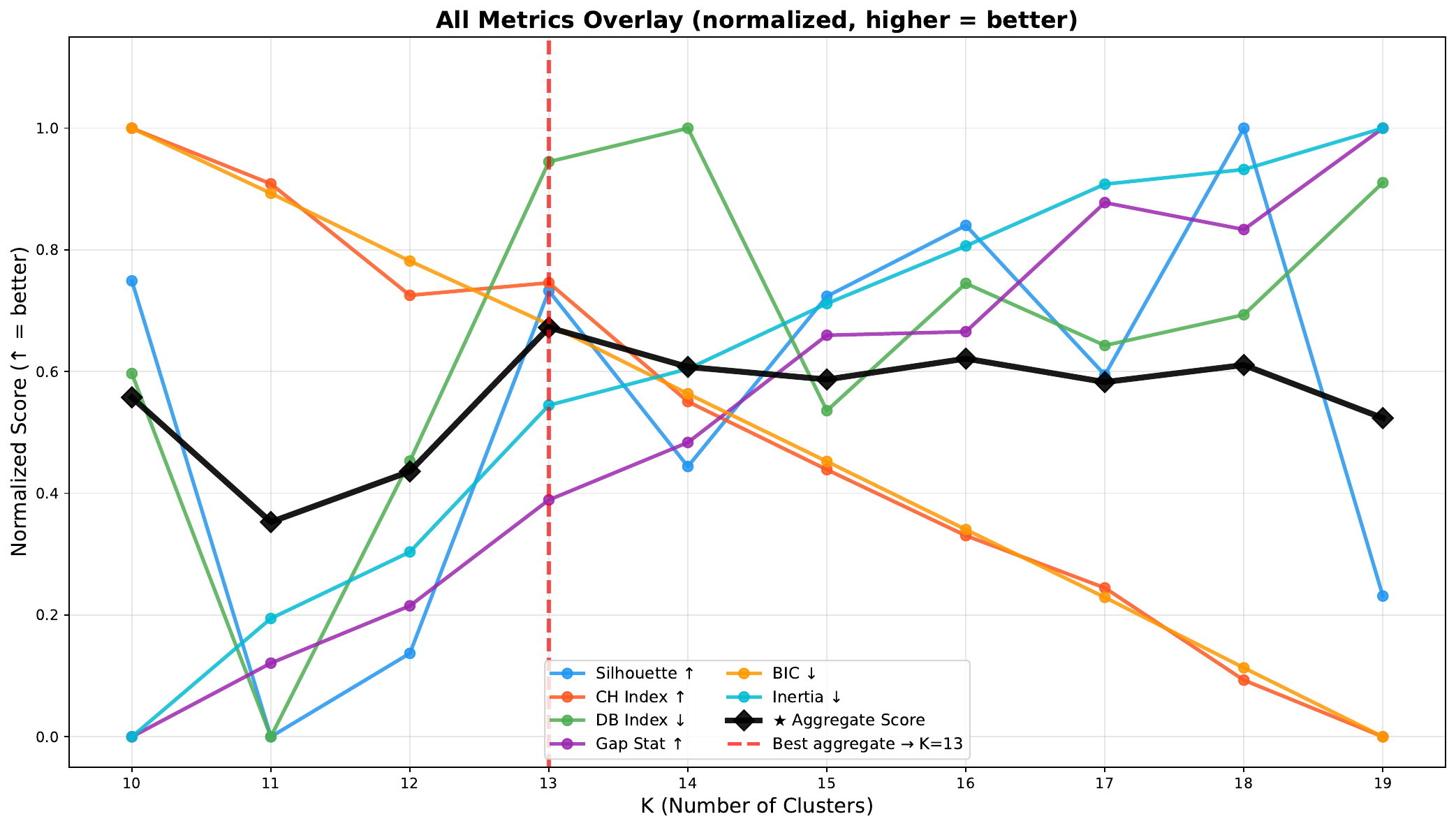}
    \end{minipage}\hfill
    \begin{minipage}{0.49\textwidth}
        \centering
        \includegraphics[width=\linewidth]{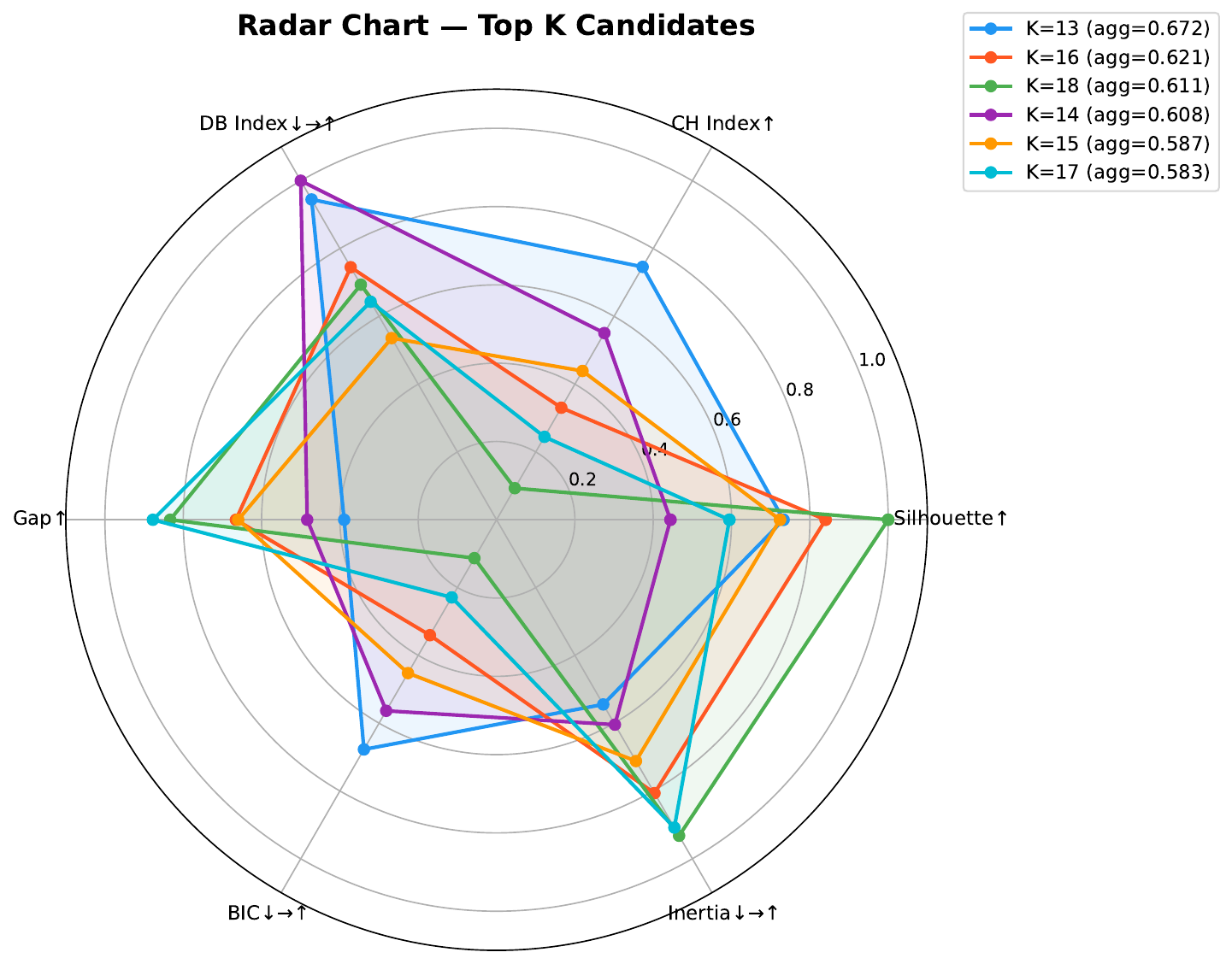}
    \end{minipage}
     \caption{\textbf{Comprehensive evaluation of clustering configurations.} (\textit{Left}) Normalized overlay of six clustering metrics across $K \in \{10, \dots, 19\}$. All metrics are scaled to $[0, 1]$ and direction-aligned so that higher values indicate better clustering quality. The black line represents the aggregate score, which peaks at $K=13$ (indicated by the red dashed line). (\textit{Right}) Radar chart comparing the multivariate performance profiles of the top-six $K$ candidates. $K=13$ achieves the highest aggregate score ($0.672$) and exhibits the most balanced, well-rounded performance across the evaluated dimensions.}
    \label{fig:kmeans}
\end{figure*}

To determine the optimal number of clusters, we conducted a systematic evaluation across \(K \in \{10, \dots, 19\}\) using six complementary clustering quality metrics: Silhouette Score, Calinski–Harabasz (CH) Index, Davies–Bouldin (DB) Index, Gap Statistic, BIC, and Inertia. To enable direct comparison, all metrics were normalized to a common \([0, 1]\) scale and directionally aligned so that higher values consistently indicate better clustering quality; we then computed an aggregate score for each \(K\). As illustrated in Fig.~\ref{fig:kmeans}, \(K=13\) achieved the highest aggregate score (0.672) and offered the best overall trade-off across metrics (radar chart). In particular, it ranks among the top three configurations in terms of Silhouette (0.061), CH (1098.6), and DB (3.103; lower is better). Notably, the CH index rebounds from \(K=12\) to \(K=13\), suggesting that the additional cluster captures meaningful structure rather than stochastic noise. This is consistent with the elbow analysis, which shows the largest single-step drop in inertia at the \(K=12 \rightarrow 13\) transition. Beyond metric performance, \(K=13\) also yields the most balanced cluster-size distribution among all configurations (coefficient of variation = 0.392; minimum cluster size = 1,754), supporting both robustness and practical interpretability.

\section{Macro-F1}
\label{sec:macro}

To complement the accuracy-focused tables in Sec.~\ref{sec:results}, this section reports Macro-F1 results under the same evaluation protocols and model groupings.

\begin{table*}[t]
\centering
\caption{Companion Macro-F1 table for Table~\ref{tab:seen-results} on seen languages (\%).}
\label{tab:appendix-seen-f1}
\begin{tabular}{lccc}
\toprule
\textbf{Model} & \textbf{13-way Macro} & \textbf{17-flat Macro} & \textbf{17-hier Macro} \\
\midrule
\multicolumn{4}{l}{\textit{Multilingual encoders}} \\
mBERT & 90.85 & 88.75 & N/A \\
XLM-R & 92.86 & 90.15 & N/A \\
\textsc{mmBERT} & 93.82 & 92.10 & N/A \\
\midrule
\multicolumn{4}{l}{\textit{Embedding-based classifiers}} \\
EmbeddingGemma & 94.12 & 92.35 & N/A \\
Gemma 3 270M (embedding+classifier) & 92.51 & 90.88 & N/A \\
\midrule
\multicolumn{4}{l}{\textit{Instruction-tuned SLMs}} \\
Gemma 3 270M & 93.22 & 91.86 & 91.95 \\
Qwen 3 0.6B & 93.55 & 92.41 & 92.56 \\
Gemma 3 1B & 94.61 & 93.12 & 93.30 \\
\midrule
\multicolumn{4}{l}{\textit{closed-source LLMs}} \\
Seed-1.8 & 92.74 & 92.57 & 91.31 \\
DeepSeek V3.2 & 93.78 & 90.33 & 88.98 \\
Qwen-Max-Preview & 95.41 & 91.48 & 93.64 \\
\bottomrule
\end{tabular}%
\end{table*}

Table~\ref{tab:appendix-seen-f1} mirrors Table~\ref{tab:seen-results} and confirms the same ranking pattern: stronger instruction-tuned models remain competitive across granularity settings, and hierarchical decoding brings modest gains for models that support structured prediction.

\begin{table*}[t]
\centering
\caption{Parent-level Macro-F1 in 17-way hierarchical evaluation for models with hierarchical decoding (\%).}
\label{tab:appendix-seen-hier-parent-f1}
\begin{tabular}{lc}
\toprule
\textbf{Model} & \textbf{17-hier Parent Macro} \\
\midrule
\multicolumn{2}{l}{\textit{Instruction-tuned SLMs}} \\
Gemma 3 270M & 93.22 \\
Qwen 3 0.6B & 93.55 \\
Gemma 3 1B & 94.61 \\
\midrule
\multicolumn{2}{l}{\textit{closed-source LLMs}} \\
Seed-1.8 & 93.63 \\
DeepSeek V3.2 & 92.51 \\
Qwen-Max-Preview & 95.54 \\
\bottomrule
\end{tabular}
\end{table*}

Table~\ref{tab:appendix-seen-hier-parent-f1} further isolates parent-level performance in hierarchical decoding. Consistent with the main-text discussion, closed-source LLMs are competitive on coarse parent routing but do not consistently surpass strong open models on long-tail robustness without task-specific training.

Finally, Table~\ref{tab:appendix-synthetic-native-f1} complements Table~\ref{tab:synthetic-native-results} by reporting Macro-F1 under matched translated-versus-native evaluation. The same overall trend holds: translated inputs are often easier than native queries, while the synthetic-to-native gap varies by model family and language.

\begin{table*}[t]
\centering
\caption{Companion F1 table for Table~\ref{tab:synthetic-native-results} under same-language translated/native evaluation. Each cell is Macro-F1 (\%).}
\label{tab:appendix-synthetic-native-f1}
\begin{tabular}{ll cccccc}
\toprule
\multirow{2}{*}{\textbf{Model}} & \multirow{2}{*}{\textbf{Lang.}} & \multicolumn{2}{c}{\textbf{13-way Parent}} & \multicolumn{2}{c}{\textbf{17-way Flat}} & \multicolumn{2}{c}{\textbf{17-way Hier}} \\
\cmidrule(lr){3-4} \cmidrule(lr){5-6} \cmidrule(lr){7-8}
& & \textbf{Trans.} & \textbf{Native} & \textbf{Trans.} & \textbf{Native} & \textbf{Trans.} & \textbf{Native} \\
\midrule
\multirow{3}{*}{mBERT}
& ZH & 57.35 & 61.89 & 61.89 & 59.93 & N/A & N/A \\
& ID & 53.56 & 52.93 & 51.78 & 50.64 & N/A & N/A \\
& TH & 17.62 & 15.66 & 18.04 & 17.93 & N/A & N/A \\
\addlinespace
\multirow{3}{*}{\textsc{XLM-R}}
& ZH & 87.13 & 82.29 & 84.74 & 80.99 & N/A & N/A \\
& ID & 90.73 & 88.85 & 88.14 & 85.96 & N/A & N/A \\
& TH & 85.68 & 88.16 & 80.67 & 80.19 & N/A & N/A \\
\addlinespace
\multirow{3}{*}{\textsc{mmBERT}}
& ZH & 90.57 & 87.20 & 87.89 & 84.54 & N/A & N/A \\
& ID & 90.66 & 88.86 & 88.37 & 85.85 & N/A & N/A \\
& TH & 84.54 & 84.02 & 82.14 & 82.67 & N/A & N/A \\
\midrule\midrule
\multirow{3}{*}{Gemma 3 270M}
& ZH & 90.85 & 88.25 & 89.75 & 86.82 & 87.05 & 87.41 \\
& ID & 88.56 & 85.93 & 86.78 & 84.14 & 87.32 & 85.58 \\
& TH & 84.15 & 79.88 & 82.81 & 81.07 & 82.76 & 79.09 \\
\addlinespace
\multirow{3}{*}{Qwen 3 0.6B}
& ZH & 92.42 & 90.64 & 91.78 & 89.89 & 91.14 & 89.99 \\
& ID & 87.22 & 85.33 & 84.85 & 82.51 & 84.35 & 82.51 \\
& TH & 83.31 & 82.07 & 83.41 & 81.72 & 82.81 & 81.37 \\
\addlinespace
\multirow{3}{*}{Gemma 3 1B}
& ZH & 94.85 & 92.27 & 94.21 & 91.92 & 94.26 & 91.13 \\
& ID & 93.71 & 91.82 & 90.39 & 88.11 & 92.52 & 90.28 \\
& TH & 88.11 & 85.49 & 86.18 & 83.85 & 87.61 & 85.18 \\
\bottomrule
\end{tabular}%
\end{table*}

\section{Training Parameters}\label{app:tp}
Experiments are conducted with both encoder-based PLMs and decoder-only LLMs. For the encoder-based baselines, we use mBERT, XLM-R, and mmBERT. These models are fine-tuned using the HuggingFace Transformers library \cite{wolf-etal-2020-transformers} with a maximum input length of 64, a batch size of 128, a learning rate of 7.0e-5, weight decay of 0.01, and max gradient norm of 5.0. For the decoder-only models, we evaluate Gemma 3 270M, Qwen 3 0.6B, and Gemma 3 1B, which are trained using LLaMA-Factory \cite{zheng2024llamafactory}. In LLM fine-tuning, we freeze all parameters except the last 16 transformer layers, and use a batch size of 128 with a learning rate of 3.0e-5. All experiments are conducted on a server equipped with 8 NVIDIA A800 GPUs.

\section{System Prompt}\label{app:systemprompt}
This section documents the prompt template used for generation-based intent classification in Sec.~\ref{sec:baselines}. The production template is XML-structured and includes role definition, strict output rules, intent definitions, and few-shot demonstrations; Fig.~\ref{fig:system-prompt} shows a representative excerpt.
\begin{figure*}[t]
\caption{Representative XML-style system prompt used for generation-based intent classification.}
\label{fig:system-prompt}
\label{sys_pro}
\begin{tcblisting}{
    colback=gray!10, % 背景颜色 (浅灰)
    colframe=gray!50, % 边框颜色 (深灰)
    arc=4pt, % 圆角大小
    listing only, 
    listing options={
        basicstyle=\ttfamily\small, % 字体和大小
        breaklines=true, % 自动换行 (非常重要，否则长 prompt 会溢出边缘)
        columns=fullflexible
    }
}
<system_prompt>
    <role>
        <title>Logistics Intent Classification Expert</title>
        <task>Analyze the user message and select exactly one intent ID.</task>
    </role>
    <core_rules description="Must be strictly observed">
        <rule id="1" name="Absolute Zero Misjudgment Principle">Only select an intent when the user's intent clearly fits the definition. If multiple intents appear, choose the dominant actionable one.</rule>
        <rule id="2" name="Output Format">Output ONLY the letter 'i' followed by the Integer ID (e.g., i7). No other words, punctuation, or explanations.</rule>
    </core_rules>
    <intent_classification_system>
        <intent id="0" name="Human Agent">
            <core_intent>User explicitly asks to talk to a real person.</core_intent>
            <definition>Request to transfer to human customer service.</definition>
            <examples>
                <example>Transfer to human customer service.</example>
                <example>I want to speak to a person.</example>
            </examples>
        </intent>
        ...
    </intent_classification_system>
    <few_shot_demonstrations>
        <demonstration>
            <user_input>My package hasn't moved for 4 days, can you urge them?</user_input>
            <output>i1</output>
        </demonstration>
        <demonstration>
            <user_input>Is it possible to ship a lighter and some fireworks?</user_input>
            <output>i10</output>
        </demonstration>
        <demonstration>
            <user_input>I want to speak to a person.</user_input>
            <output>i0</output>
        </demonstration>
    </few_shot_demonstrations>
</system_prompt>
\end{tcblisting}
\end{figure*}

\end{document}